\documentclass[journal]{IEEEtran}
\IEEEoverridecommandlockouts
\usepackage[caption=false,font=normalsize,labelfont=sf,textfont=sf]{subfig}
\usepackage{cite}
\usepackage{amsmath,amssymb,amsfonts}
\usepackage{algorithm}
\usepackage{algpseudocode}
\usepackage{graphicx}
\usepackage{textcomp}
\usepackage{xcolor}
\def\BibTeX{{\rm B\kern-.05em{\sc i\kern-.025em b}\kern-.08em
    T\kern-.1667em\lower.7ex\hbox{E}\kern-.125emX}}
\usepackage{hyperref}
\usepackage{soul}
\usepackage{float}
\usepackage{bm}
\usepackage{booktabs} % For professional table lines
\usepackage{array} % Better column alignment
\usepackage{multirow}
\usepackage{url}
\sethlcolor{yellow}

\makeatletter
\@ifundefined{IEEEImpStatement}{%
  \newenvironment{IEEEImpStatement}[1][Impact Statement]{%
    \par\addvspace{1.5\baselineskip}%
    \noindent{\bfseries #1}\par\vspace{0.4\baselineskip}%
    \noindent\ignorespaces
  }{%
    \par\addvspace{1.0\baselineskip}%
  }%
}{}
\makeatother

\begin{document}
\title{From Graphs to Gradients: Physics-Inspired Structural Attribution for Cyber-Physical IoT Systems and Beyond}
\author{Spyridon Evangelatos\thanks{S.~Evangelatos is with the Research \& Innovation Development Department, Netcompany S.A.,  Luxembourg (email: sevangelatos@netcompany.com).}, Christos Diou, \IEEEmembership{Member, IEEE}\thanks{C.~Diou and G.~Papadopoulos are with the Department of Informatics and Telematics, Harokopio University of Athens, Greece, (emails: cdiou@hua.gr and g.th.papadopoulos@hua.gr).}, Georgios Th. Papadopoulos, \IEEEmembership{Member, IEEE}, Evangelos Markakis,~\IEEEmembership{Member,~IEEE}\thanks{E. Markakis is with the  Department of Electrical and Computer Engineering, Hellenic Mediterranean University, Greece, (email: emarkakis@hmu.gr).}, and Panagiotis Sarigiannidis, \IEEEmembership{Member, IEEE}\thanks{P.~Sarigiannidis is with the Department of Electrical and Computer Engineering, University of Western Macedonia, Greece, (email: psarigiannidis@uowm.gr).}}

%\markboth{Journal of IEEE Transactions on Artificial Intelligence, Vol. 00, No. 0, Month 2026}%
%{S.~Evangelatos \MakeLowercase{\textit{et al.}}: Physics-Inspired Causal Explanation for Cyber-Physical IoT Systems}
%\IEEEpubid{0000--0000/00\$00.00~\copyright~2026 IEEE}

\maketitle

\begin{abstract}
Interpretable explanation methods in Artificial Intelligence aim to uncover the underlying causes and their effects, enabling a deeper understanding of why a system behaves in a certain way under different inputs. In contrast to traditional explainability methods, which mainly highlight correlations between input variables and their corresponding output variables, causal explanation focuses on answering interventional questions. By doing so, it provides more robust insights, assisting users in better understanding automated decisions, especially in high-risk domains.

Recovering an explicit directed causal structure, however, is often impractical in large-scale, hybrid cyber-physical systems characterised by feedback loops and partial observability. This paper introduces a novel framework inspired by concepts drawn from the field of statistical mechanics that instead models variable dependencies through an undirected, energy-based representation of cyber-physical IoT systems. Our approach enables rigorous dependency-aware attribution by analysing how variations in the energy landscape reflect the influence of individual components within the system, without requiring the recovery of a directed causal graph. In addition, it also supports reasoning about the effects of perturbations across hybrid interactions, providing reliable explanations of abnormal behaviours.

Our framework was empirically examined through extensive simulations on an industrial IoT testbed that consists of hybrid continuous and discrete variables, demonstrating higher attribution accuracy, improved robustness to perturbations and substantially better scalability compared to state-of-the-art graph-based explainability approaches. While the resulting attributions are not intended to fully recover or explain the system's underlying generative dynamics, they provide valuable, dependency-aware explanations that support both human interpretation and downstream predictive and diagnostic tasks. Although it is demonstrated in industrial IoT security, our framework is also applicable to other high-dimensional cyber-physical and socio-technical systems that require principled, structural explanations.
\end{abstract}

\begin{IEEEImpStatement}
Modern cyber-physical systems, such as industrial IoT infrastructures and water treatment plants, are increasingly exposed to complex and often hidden failure mechanisms. When incidents occur, operators need clear explanations of what caused them and how they propagate across interconnected components. The framework introduced in this paper provides such explanations without relying on hard-to-obtain directed causal graphs or opaque model-specific techniques, making it suitable for systems with many interacting variables.

Experimental results on the \texttt{SWaT} testbed show that the proposed method approaches perfect identification of root causes, while competing approaches remain close to zero. In addition, our method maintains higher attribution accuracy and robustness as the system dimensionality increases. These improvements can be translated into faster and more reliable fault diagnosis in critical infrastructure, helping reduce downtime, safety risks and operational costs. More broadly, our work contributes to trustworthy AI by offering a scalable and interpretable approach to system-level attribution in complex, real-world environments.
\end{IEEEImpStatement}


\section{Introduction}
\IEEEPARstart{A}{s} Artificial Intelligence (AI) systems increasingly influence critical decision-making processes, the demand for transparency and interpretability has become of paramount importance. Traditional explainability methods often rely on correlational associations, which offer limited insight into how different variables influence a model's behaviour under changing conditions. Causal explainability~\cite{Moraffah:2020} aims to bridge this gap by examining how perturbations or interventions on specific variables may influence the resulting system behaviour.

This perspective is grounded in causal inference theory~\cite{Jonas:2017} and often formalised using tools such as structural causal models and graphical representations. Nevertheless, these classical inference frameworks typically rely on predefined directed structures, which are frequently difficult or impossible to identify reliably from observational data in systems exhibiting feedback loops and partial observability. As a result they struggle to scale to large, hybrid cyber-physical systems characterised by feedback loops, mixed continuous-discrete variables, and partially observable dynamics. They are also sensitive to model misspecification and distributional shifts, while remaining difficult to scale in adversarial or abnormal operating regimes.

To address these limitations, alternative formulations are required that can capture complex dependencies without imposing rigid causal hierarchies. At the same time, such formulations must remain scalable and interpretable in high-dimensional settings. In cyber-physical IoT systems, system-level behaviour often emerges from local interactions among heterogeneous components rather than from explicitly defined causal chains. Probabilistic representations are therefore needed to capture both normal and abnormal operating regimes.

In this paper, we propose an energy-based explanation framework inspired by principles from statistical mechanics. Rather than positing a directed causal structure, the system is modelled through an undirected energy landscape that assigns low energy values to configurations that are consistent with normal system behaviour and higher energy values to abnormal or adversarial states. Attribution scores are obtained by analysing how local perturbations, global free-energy variations and higher-order interactions reshape the overall energy landscape. In this way, our framework supports both local and global analysis of complex hybrid cyber-physical systems under perturbed operating conditions, without requiring the explicit recovery of a directed causal graph. The resulting attributions are valuable for explanation and for downstream prediction and detection tasks, but, unlike a fully specified directed causal model, they are not intended to explain the complete generative dynamics of the system.  
\IEEEpubidadjcol

The proposed framework is evaluated on a large scale industrial IoT testbed, where both controlled perturbations and realistic attack scenarios are introduced on hybrid cyber-physical data. Its performance is assessed in terms of attribution accuracy, robustness to input perturbations and computational scalability. The results demonstrate consistent improvements over well established explainability methods based on graph theory,  particularly as the size and complexity of the system increase.

The key contributions of this paper are summarised below:
\begin{itemize}
  \item the formulation of an energy-based dependency-aware attribution framework, grounded in an undirected graphical representation, that provides interpretable explanations for hybrid cyber-physical systems;
  \item a thorough evaluation of the framework conducted on the \texttt{SWaT} industrial testbed~\cite{Mathur:2016} that demonstrates accurate identification of dominant contributing variables and their downstream effects during a realistic cyber-physical attack; and
\item a comparison showing that our framework achieves higher explanatory accuracy and significantly better computational scalability than GNNExplainer across increasing system sizes.
\end{itemize}

For ease of reference, Table~\ref{tab:Notat} summarises the key notation used throughout this paper, revealing the connection between the probabilistic energy based modelling and concepts from statistical mechanics.

The rest of the paper is organised as follows: Section~\ref{Sec:RelaWo} provides a literature review on the related work on explainability and interpretable AI approaches for cyber-physical IoT systems. Section~\ref{Sec:FraDes} describes in detail the proposed energy-based attribution framework, formalising the underlying probabilistic modelling of the system, along with the local and global attribution mechanisms and the associated computational pipeline. Section~\ref{Sec:IoTSec} describes the application of the framework to IoT cybersecurity scenarios, with emphasis on industrial cyber-physical systems. Section~\ref{Sec:ResAna} provides an extensive experimental evaluation, such as attribution accuracy, robustness to perturbations and scalability comparisons against state-of-the-art methods. Finally, Section~\ref{Sec:ConFut} concludes the paper and indicates future research directions.

\section{Related Work}\label{Sec:RelaWo}
Causal explainability in AI has evolved through complementary theoretical frameworks and practical methods~\cite{Rawal:2025}, including structural causal models, counterfactual explanation methods, probabilistic graphical models, graph-based explainability approaches and physics-inspired and information-theoretic techniques.

Structural causal models (SCM) provide a principled framework for representing cause-effect relationships through explicit variables and directed dependencies. The authors of~\cite{Huang:2020} propose a causal discovery framework that exploits distributional changes across heterogeneous and non-stationary environments to identify invariant causal mechanisms and recover causal structures from observational data. However, their approach focuses on causal graph identification and relies on explicit structural assumptions. This makes it less suitable for scalable causal attribution and explanation in large, hybrid cyber-physical systems with complex feedback and interaction dynamics. In addition, the approach can also incur significant computational overhead as the number of variables and environments increases.

\begin{table}[!t]
\centering
\renewcommand{\arraystretch}{1.1}
\caption{\textbf{Notation Summary}}
\vspace{-5pt}
\begin{tabular}{@{}ll@{}}
\toprule
$\mathbf{x} $ & Joint state vector of the system variables \\
$\mathbf{x}_{-i}$ & Joint configuration of all system variables except $x_{i}$\\
$\mathcal{X}_{i}$ & Domain of variable $x_{i}$ (continuous or binary) \\
$\mathcal{X} = \prod_{i=1}^{n} \mathcal{X}_i$ & Joint configuration space \\
$\mathcal{G} = (\mathcal{V}, \mathcal{D})$ & Undirected graphical model with nodes $\mathcal{V}$ and edges $\mathcal{D}$ \\
$\mathcal{E}(\mathbf{x})$ & Energy function associated with configuration $\mathbf{x}$ \\
$\phi_i(x_{i})$ & Unary potential function for variable $x_{i}$ \\
$\phi_{ij}(x_{i}, x_{j})$ & Pairwise potential function between $x_{i}$ and $x_j$ \\
$\beta$ & Inverse temperature parameter \\
$\mathcal{Z}$ & Partition function ensuring normalisation \\
$p(\mathbf{x})$ & Boltzmann distribution over system states \\
$\Delta_i$ & First-order gradient-based sensitivity score \\
$\mathcal{F}(x_{i})$ & Free energy conditioned on $x_{i}$ \\
$\mathcal{H}(\mathbf{x}_{-i}|x_{i})$ & Conditional entropy given $x_{i}$ \\
$\gamma_i$ & Second-order curvature score for variable $x_{i}$ \\
$\gamma_{ij}$ & Interaction score between variables $x_{i}$ and $x_{j}$ \\
$M$ & Number of Monte Carlo samples used for estimation \\
$k$ & Number of top-ranked variables selected for refinement\\
\bottomrule
\end{tabular}\label{tab:Notat}
\end{table}

In~\cite{Zeng:2025}, the authors introduce a causal anomaly detection method that reduces misleading correlations and provide interpretable explanations through structural causal model pruning and counterfactual reasoning. The approach demonstrates consistent accuracy gains across heterogeneous IoT datasets while enabling root-cause analysis. However, the higher computational cost and the dependence of the method on accurate causal modelling may hinder scalability and degrade performance when underlying causal relationships evolve or are only partially observed.

Counterfactual explanations aim to explain model decisions by identifying minimal changes to the input that would lead to a different outcome~\cite{Verma:2024}. In this context, the authors of~\cite{Karimi:2021} reformulate algorithmic recourse by shifting from nearest counterfactual explanations to minimal structural interventions. These interventions consider explicitly real-world causal dependencies among features. The authors show that naive counterfactual-based recommendations fail when features are causally related. To address this, they propose a causal framework based on structural causal models and the abduction-action-prediction pipeline to generate feasible recourse actions that respect downstream effects. 

Nevertheless, their approach assumes full knowledge of the structural causal model, a restrictive requirement that is rarely satisfied in practice. It further relies on additive noise models with invertible structural equations, limiting applicability to complex hybrid systems with feedback and mixed continuous-discrete dynamics~\cite{Zhang:2022}. Additionally, the framework focuses on recourse optimisation in individual level. It does not address the challenge of discovering or learning causal structures from data when the true causal model is unknown, relying instead on pre-specified graphs. 

Related concerns are also highlighted by~\cite{Bhatt:2021}, which emphasises the challenges of deploying explainable and causal machine learning methods under real-world operational constraints. In particular, classical counterfactual approaches commonly assume independent features, an assumption that is violated in tightly coupled cyber-physical environments.

Probabilistic graphical models capture causal and statistical dependencies among variables through structured probabilistic representations. In this context, the authors of~\cite{Gad:2025} introduce the TOCA-IoT framework which combines causal discovery based on LiNGAM~\cite{Shimizu:2006} with explainable machine learning techniques. LiNGAM is a probabilistic graphical approach used to identify directed acyclic graphs that capture causal relationships. Within TOCA-IoT, this combination is used to model directed dependencies among IoT network features, supporting more interpretable anomaly detection. The resulting causal graph is then used to support interpretable reasoning about the root causes of network attacks, while also enhancing detection performance through threshold optimisation. However, this approach relies on linear, acyclic causal assumptions inherent to LiNGAM, which limits its ability to represent feedback loops and non-linear interactions often met in cyber-physical systems. In parallel, causal structure learning is performed offline on static datasets, which restricts adaptability to evolving system dynamics and online inference scenarios.

Explainability methods for graph-structured models aim to interpret predictions by identifying influential nodes, edges and substructures within a graph. These approaches primarily explain the behaviour of trained predictive models, rather than the underlying system dynamics that generate the data. In their seminal work in~\cite{Ying:2019:GNNE}, the authors introduce GNNExplainer, a model-agnostic framework for explaining the predictions of graph neural networks. Their method identifies a compact subgraph and a subset of node features that are most influential for a given prediction. It formulates explanation as an optimisation problem that maximises mutual information between the explanation and the model’s output, enabling instance level interpretability for graph structured data. Despite its effectiveness, the approach explains the behaviour of a trained predictive model rather than the underlying system or data-generating process, limiting causal interpretability. Moreover, its optimisation-based explanation procedure can be computationally expensive and sensitive to noise, restricting scalability and robustness in large, dynamic cyber–physical systems.

Complementary to model-centric and graph-based explanation methods, physics-inspired and information-theoretic approaches frame explainability in terms of fundamental principles such as energy and entropy. In a recent work~\cite{Mehdi:2024}, the authors introduce a thermodynamics-inspired, model-agnostic explainability framework that formulates explanations as an energy-entropy trade-off. It uses concepts such as free energy and interpretation entropy to quantify human interpretability. The approach grounds explanations in information-theoretic and thermodynamic principles. This provides a coherent way to balance fidelity and simplicity while offering insights into the factors driving model predictions. Nevertheless, the method is primarily designed to explain black-box predictive models. It does not explicitly model system-level causal interactions and relies on local surrogate models and neighbourhood sampling, which limits its ability to capture global broader dependencies and dynamic feedback in large-scale cyber-physical systems.

In a recent work~\cite{Evangelatos:2025}, the authors introduce an energy-based formulation for generating counterfactual explanations by modelling decision boundaries through probabilistic energy landscapes. The proposed framework draw on concepts such as energy minimisation and local landscape perturbations, providing a rigorous mechanism for the identification of minimal counterfactual changes while also accounting for uncertainty. Despite its effectiveness, this framework primarily targets counterfactual explanations at the decision level and does not explicitly capture how interactions between system components influence the overall system behaviour. On top of that, causal influence is assessed locally, without capturing global dependencies that arise in complex, large scale cyber-physical systems.

Compared to existing approaches, our framework avoids the need to recover an explicit directed causal graph, instead modelling variable dependencies through an undirected probabilistic energy landscape that scales to large hybrid systems, and provides a unified attribution mechanism. Structural causal models and causal discovery techniques typically depend on predefined directed graphs and strong structural assumptions, which can limit their applicability in settings with feedback loops, mixed continuous-discrete dynamics and partial observability. It is important to emphasise that, in this work, causality is interpreted through the effect that the perturbations of the system variables have on the overall system behaviour, rather than through the explicit recovery of directed causal structures from observational data alone.

Counterfactual and recourse-based methods typically focus on decision-level explanations and thus, do not capture system-wide interactions arising from interactions among components. Probabilistic graphical models and graph-based explainability techniques primarily interpret the behaviour of trained predictive models, rather than the underlying system dynamics, and often require labelled data or costly optimisation procedures. In contrast, our framework builds upon principles from statistical mechanics and information theory in order to provide interpretable, system-level explanations that remain robust under noise and adversarial conditions in large-scale IoT environments.

\section{Framework Description}\label{Sec:FraDes}
Let $\mathbf{x} = (x_{1}, x_{2}, \dots, x_{n})$ represent the state of a cyber-physical system, where each variable $x_{i}\in\mathcal{X}_{i}$ corresponds to a measurable component of the system. The domain $\mathcal{X}_i$ may be continuous, $\mathcal{X}_i \subseteq \mathbb{R}$ for variables such as temperature or network throughput or binary, $\mathcal{X}_i = {0,1}$, for actuator states, alerts, sensor triggers, etc. The joint configuration space is defined as
\begin{equation}
\mathcal{X} = \prod\limits_{i=1}^{n} \mathcal{X}_{i}
\end{equation}
forming a hybrid domain that includes both continuous and binary components. This structure allows the model to handle heterogeneous data types within a unified probabilistic framework.

Our goal is to model the probabilistic structure of an IoT system in order to identify which variables contribute most strongly to abnormal cybersecurity events. This requires a joint distribution that can accurately capture both typical operational patterns and deviations that may indicate anomalous or malicious activity, which we express using a Boltzmann distribution~\cite{Jaynes:1957}
\begin{equation}
p(\mathbf{x}) = \frac{1}{\mathcal{Z}} \exp\left( -\beta \mathcal{E}(\mathbf{x}) \right).
\end{equation}
In this formulation, $\mathcal{E}(\mathbf{x})$ represents the energy function that captures the system’s internal dependencies.The function operates by assigning lower energy values to to system states that are more consistent with normal behaviour and higher energy values to those that deviate from expected patterns. The inverse temperature parameter $\beta$ controls the sharpness of the distribution by adjusting the relative importance between high- and low-energy states. The partition function~\cite{Potamianos:1997:PartitionFun}, $\mathcal{Z}$, defined as
\begin{equation}
\mathcal{Z} = \sum \int\exp\left[-\beta \mathcal{E}(\mathbf{x})\right]\mathrm{d}\mathbf{x},
\end{equation}
ensures proper normalisation, where summation and integration operations address the discrete and continuous state components of $\mathbf{x}$ respectively .

Instead of relying on a directed acyclic graph, which not only restricts the causal structure to hierarchical relationships but is also frequently unidentifiable or unrecoverable from observational data in systems with feedback loops and partial observability, we adopt a more flexible, undirected representation. Specifically, we model the system's structure as a graphical model $\mathcal{G} = (\mathcal{V}, \mathcal{D})$, where $\mathcal{V}$ denotes the set of variables and $\mathcal{D}\subseteq \mathcal{V}\times\mathcal{V}$ the set of edges representing statistical or functional dependencies among variables. A key advantage of this representation is its ability to capture complex interactions, including feedback loops and symmetric relationships common in cyber-physical systems. 

The local structure of the graph enables the energy function $\mathcal{E}(\mathbf{x})$ to be decomposed as
\begin{equation}
\mathcal{E}(\mathbf{x}) = \sum\limits_{(i,j)\in\mathcal{D}}\phi_{ij}(x_{i}, x_{j}) + \sum\limits_{i\in\mathcal{V}}\phi_{i}(x_{i}).
\end{equation}
The pairwise potential functions $\phi_{ij}(x_{i}, x_{j})$ capture the interaction energy among the connected variable pairs $(x_{i}, x_{j})$, while the unary potentials $\phi_{i}(x_{i})$ reflect intrinsic preferences or constraints governing individual variables. The choice of the functional form depends on the nature of the variables. Continuous variables can be modelled using quadratic potentials, whereas binary interactions can be represented using logistic or indicator-based functions, depending on the application. 

The above mentioned factorisation allows the overall system behaviour to be modelled in terms of local dependencies, enabling tractable probabilistic inference and interpretable system-level analysis in high-dimensional settings. The proposed graphical representation is not intended to explicitly recover directed causal structures from observational data alone. Instead, it is designed to capture interaction patterns and quantify how perturbations propagate through the system under different operating conditions.

Within IoT-based cyber-physical operational contexts, low-energy configurations correspond to the system states that reflect typical, expected operational patterns, such as consistent actuator behaviour and plausible network activity. In contrast, configurations that involve conflicting device states are assigned higher energy, reflecting their deviation from the system’s normal operating conditions.

\subsection{Local Causal Attribution via Energy Sensitivity}
In cybersecurity settings, understanding why a particular alert was triggered given the observed system variables is critical. Our energy-based probabilistic framework enables this form of explanation through the quantification of the influence of each variable on the system's energy landscape. More specifically, we focus on computing the local sensitivity scores that capture how changes in a variable affect the energy associated with an observed configuration.

Given an observed system state $\mathbf{x}$, we consider the partial derivative of the energy function with respect to a particular variable $x_{i}$, i.e., 
\begin{equation}
\Delta_{i} = \Bigg| \frac{\partial\mathcal{E}(\mathbf{x})}{\partial x_{i}} \Bigg|.
\end{equation}
These scores should be interpreted as local measures of how sensitive the energy landscape is to perturbations in individual variables, rather than as direct estimates of causal effects. The motivation of this formulation lies in the relationship, embedded in the Boltzmann distribution, between a configuration’s energy and its probability. In regions where the energy landscape changes sharply with respect to certain variables, even small perturbations can lead to noticeable changes in the likelihood of the system’s state. Hence, the magnitude of the energy gradient with respect to $x_{i}$ provides an indication of how strongly that variable contributes to the system’s behaviour in the observed configuration.

The gradient magnitude serves as a local attribution score: variables for which $\Delta_{i}$ is large contribute significantly to the total energy, and thus play a strong explanatory role in determining the likelihood of the observed state. Intuitively, a variable with a high gradient is one for which small perturbations would induce a large change in energy, suggesting that it is critical to the maintenance or disruption of the system’s equilibrium.

When the energy function is factorised over a graphical model, we can exploit locality to compute gradients more efficiently.  In our case, where the variable $x_{i}$ is connected only to a small number of neighbours $x_{j}$ then 
\begin{equation}\label{eqn:EnerFirsDer}
\frac{\partial\mathcal{E}(\mathbf{x})}{\partial x_{i}} = \sum\limits_{j:(i,j)\in\mathcal{D}}\frac{\partial \phi_{ij}(x_{i}, x_{j})}{\partial x_{i}} + \frac{\partial \phi_{i}(x_{i})}{\partial x_{i}}.
\end{equation}

The resulting expression demonstrates how causal attributions remain local in the graph structure and are computationally tractable, especially in sparse graphs. For binary variables, we approximate derivatives using energy differences, i.e.,
\begin{equation}
\Delta_{i} = \Big| \mathcal{E}(\mathbf{x}_{-i}, x_{i}=1) -  \mathcal{E}(\mathbf{x}_{-i}, x_{i}=0) \Big|,
\end{equation}
where $\mathbf{x}_{-i}$ denotes the vector $\mathbf{x}$ with all its entries unchanged except for the $i$-th, which is explicitly varied to compute the energy difference. This local sensitivity-based attribution is particularly well suited to real-time or explainable cybersecurity systems, where understanding the root cause of an anomaly must be both fast and interpretable. 

\subsection{Global Attribution via Free Energy}
While gradient-based sensitivity offers a local explanation of causal relevance with respect to a specific system state, it does not capture how consistently a variable contributes to outcomes across the full distribution of possible configurations. To overcome this limitation and move beyond local gradient-based explanations, we introduce a complementary attribution method based on the concepts of entropy and free energy from statistical physics. This approach allows us to assess which variables consistently influence the probability of alert events across the full distribution, instead of focusing only on a single configuration.

The free energy associated with a configuration conditioned upon $x_{i}$ is formulated as
\begin{equation}\label{eqn:FreeEner}
\mathcal{F}(x_{i}) = -\frac{1}{\beta}\log\left[\sum_{\mathbf{x}_{-i}} \exp\left(-\beta \mathcal{E}(x_{i}, \mathbf{x}_{-i})\right)\right],
\end{equation}
which can be rewritten in the standard form that separates energetic and entropic contributions, i.e., 
\begin{equation}
\mathcal{F}(x_{i}) = \mathbb{E}_{\mathbf{x}_{-i}|x_{i}}\left[\mathcal{E}(x_{i}, \mathbf{x}_{-i}) - \frac{1}{\beta}\mathcal{H}(\mathbf{x}_{-i}|x_{i})\right], 
\end{equation}
where $\mathcal{H}(\mathbf{x}_{-i}|x_{i})$ denotes the conditional entropy of the remaining variables given $x_{i}$. The first term in Eq.~\eqref{eqn:FreeEner} corresponds to the average internal energy under this conditional distribution, while the second term captures the level of uncertainty once $x_{i}$ is fixed. If assigning a value to $x_{i}$ reduces both the expected energy and the entropy, then the system is effectively driven into a more stable and predictable regime. This indicates that $x_{i}$ is highly informative and possibly a causal driver of the configuration.

To assess the influence of a variable beyond a single observation, we consider the expected free energy taken with respect to the marginal distribution of $x_{i}$, with the expression depending on whether the variable is discrete or continuous. For discrete variables, the expectation becomes
\begin{equation}
\mathbb{E}[\mathcal{F}(x_{i})] = \sum\limits_{x_{i}} p(x_{i}) \mathcal{F}(x_{i}),
\end{equation}
whereas for continuous variables, we compute
\begin{equation}
\mathbb{E}[\mathcal{F}(x_{i})] = \int p(x_{i}) \mathcal{F}(x_{i}) \mathrm{d}x_{i}.
\end{equation}
In both cases, these expectations indicate how conditioning on a particular variable value reshapes the overall distribution and, therefore, how much that variable contributes to the system’s global behaviour. In In settings such as cyber-physical IoT infrastructures, which mix binary actuator states with continuous sensor readings, this approach offers a coherent way to assess global causal attribution based on thermodynamic principles.

\subsection{Second-Order Attribution via Curvature}
Despite the fact that the first-order derivatives of the energy function in Eq.~\eqref{eqn:EnerFirsDer} provide valuable information about how sensitive the system's energy is to small perturbations in individual variables, they do not fully capture the nature of variable interactions or the stability of these influences. In order to address this, we incorporate second-order derivatives, namely, the diagonal and off-diagonal entries of the Hessian matrix of the energy function, as a refinement of our attribution framework. The inclusion of these curvature-based metrics allows us to measure not just how much a variable influences the energy, but also how confidently and consistently that influence behaves in the local energy landscape.

Formally, the second-order structure of the energy function is captured by its Hessian matrix, denoted by
\begin{equation}
\mathbf{H}_{\mathcal{E}(\mathbf{x})}(\mathbf{x}) = \left[ \frac{\partial^{2} \mathcal{E}(\mathbf{x})}{\partial x_{i} \partial x_{j}} \right]_{i,j=1}^{n},
\end{equation}
which is a symmetric $n \times n$ matrix encoding both the local curvature along each variable (diagonal terms) and the pairwise interactions (off-diagonal terms). For a given variable $x_{i}$, the diagonal entry $\frac{\partial^{2} \mathcal{E}(\mathbf{x})}{\partial x_{i}^{2}}$ quantifies the convexity of the energy landscape in that direction, while $\frac{\partial^{2} \mathcal{E}(\mathbf{x})}{\partial x_{i} \partial x_{j}}$ describes the second-order coupling between $x_{i}$ and $x_{j}$. In other words, mixed partial derivatives capture the interaction effects between variables, revealing whether the causal impact of one variable depends on the state of another. In practice, when the energy function is factorized over a graph, most off-diagonal terms vanish due to sparsity, and the non-zero entries of the Hessian matrix correspond only to directly connected variable pairs.

Beyond curvature second-order information can also be interpreted from an entropy-informed perspective. When considering the energy landscape defined over $\mathbf{x}_{-i}$ conditioned on $x_{i}$,  the local curvature relates to the sharpness of the conditional distribution $p(\mathbf{x}_{-i}|x_{i})$ A sharper distribution implies lower entropy and thus a more deterministic relationship between  and the rest of the system. This connection can be made explicit by approximating the conditional entropy using the curvature of the log-density:
\begin{equation}\label{eqn:curvature}
\mathcal{H}(\mathbf{x}_{-i}|x_{i}) \approx \frac{1}{2}\log\left[(2\pi e)^{k}\det\left[\beta\nabla_{\mathbf{x}_{-i}}^{2}\mathcal{E}(x_{i}, \mathbf{x}_{-i})\right]^{-1}\right],
\end{equation}
where $k$ is the dimensionality of $\mathbf{x}_{-i}$ and the Hessian is evaluated at the mode or mean of the distribution. This approximation, valid for nearly Gaussian conditional posteriors, illustrates how entropy and curvature are inherently connected and how the second-order structure of the energy function contributes to the informational profile of the model. 

Although second-order information increases expressiveness, computing the full Hessian matrix is computationally demanding, especially in large-scale systems. In addition, second-order derivatives are typically more sensitive to noise, particularly in sparsely sampled domains. In order to balance interpretability and efficiency, we adopt a selective approach where the second-order derivatives are computed only for the top-ranked variables identified by their first-order sensitivity scores. This strategy, allow us to focus the computational effort on the most influential regions of the system state space, refining our explanations without incurring the full cost of dense second-order analysis.

To quantify these effects in practice, we define a scalar curvature-based attribution score for each variable $x_{i}$ as:
\begin{equation}
\gamma_{i} = \Big|\frac{\partial^{2} \mathcal{E}(\mathbf{x})}{\partial x_{i}^{2}}\Big|,
\end{equation}
which measures the degree to which the energy landscape curves around $x_{i}$, capturing the stability or volatility of its influence. Furthermore, we introduce a second-order interaction score for pairs of variables:
\begin{equation}
\gamma_{ij} = \Big|\frac{\partial^{2} \mathcal{E}(\mathbf{x})}{\partial x_{i}\partial x_{j}}\Big|,
\end{equation}
which reflects how strongly the explanatory effect of $x_{i}$ depends on the state of $x_{j}$, and vice versa. These values are subcomponents of the full Hessian matrix and provide interpretable scalar metrics for attribution ranking. The scores can be aggregated over all neighbors in the graphical model to obtain a refined structural profile of causal interdependence.

\begin{algorithm}[!t]
\caption{Causal Attribution via Energy-Based Analysis}
\begin{algorithmic}[1]
\State \textbf{Input:} Observed configuration $\mathbf{x}$, graphical model $\mathcal{G}=(\mathcal{V}, \mathcal{D})$, energy function $\mathcal{E}(\cdot)$, potential functions $\phi_i(\cdot)$, $\phi_{ij}(\cdot, \cdot)$, inverse temperature $\beta$, number of top-ranked variables $k$.
\State \textbf{Output:} Attribution scores $\Delta_i$, $\mathcal{F}(x_{i})$, $\gamma_i$, $\gamma_{ij}$
\For{each variable $x_{i} \in \mathcal{V}$}
\State Compute local energy sensitivity:
\State $\Delta_{i} \gets \left| \frac{\partial \mathcal{E}(\mathbf{x})}{\partial x_{i}} \right|$ or finite-difference for binary $x_{i}$
\EndFor
\For{each variable $x_{i} \in \mathcal{V}$}
\State Initialize $\mathbf{x}_{-i}^{(0)}$ randomly
\For{$m = 1$ to $M$}
\State Propose a new state $\mathbf{x}_{-i}'$ using symmetric proposal distribution $q\left(\mathbf{x}_{-i}' | \mathbf{x}_{-i}^{(m-1)}\right)$
\State Compute acceptance probability:
\State \hspace{-3em} $\alpha = \min\left(1, \exp\left[-\beta\left(\mathcal{E}(x_{i}, \mathbf{x}_{-i}') - \mathcal{E}\left(x_{i}, \mathbf{x}_{-i}^{(m-1)}\right)\right)\right]\right)$
\State Accept or reject proposal:
\State $\mathbf{x}_{-i}^{(m)} \gets \mathbf{x}_{-i}'$ with probability $\alpha$, 
\State else $\mathbf{x}_{-i}^{(m)} \gets \mathbf{x}_{-i}^{(m-1)}$
\EndFor
\State Estimate expected energy:
\State \hspace{1em} $\mathbb{E}[\mathcal{E}] \gets \frac{1}{M} \sum\limits_{m=1}^{M} \mathcal{E}(x_{i}, \mathbf{x}_{-i}^{(m)})$
\State Estimate conditional entropy:
\State \hspace{1em} $\mathcal{H}(\mathbf{x}_{-i}|x_{i}) \gets -\frac{1}{M} \sum\limits_{m=1}^{M} \log \left[p\left(\mathbf{x}_{-i}^{(m)} | x_{i}\right)\right]$
\State Compute free energy:
\State \hspace{1em} $\mathcal{F}(x_{i}) \gets \mathbb{E}[\mathcal{E}] - \frac{1}{\beta} \mathcal{H}(\mathbf{x}_{-i}|x_{i})$
\EndFor
\State Rank variables by $\Delta_{i}$ or $\mathcal{F}(x_{i})$ and select top-$k$ influential ones
\For{each top-$k$ variable $x_{i}$}
\For{each neighbor $x_j$ such that $(i,j)\in\mathcal{D}$}
\State Compute second-order curvature:
\State \hspace{1em} $\gamma_i \gets \left| \frac{\partial^2 \mathcal{E}}{\partial x_{i}^2} \right|$, \quad $\gamma_{ij} \gets \left| \frac{\partial^2 \mathcal{E}}{\partial x_{i} \partial x_j} \right|$
\EndFor
\EndFor
\State \Return ${\Delta_i}, {\mathcal{F}(x_{i})}, {\gamma_i}, {\gamma_{ij}}$
\end{algorithmic}\label{Alg:Algo}
\end{algorithm}

\subsection{Monte Carlo Estimation of Energy and Entropy Contributions}
To overcome the computational intractability of evaluating conditional expectations and entropies in high-dimensional hybrid domains, we employ Monte Carlo techniques~\cite{Browne:2012} to approximate these quantities. In particular, we estimate the expected energy and the conditional entropy used in the free energy decomposition through samples drawn from the conditional distribution $p(\mathbf{x}_{-i}| x_{i})$, which itself inherits a Boltzmann structure:
\begin{equation}
p(\mathbf{x}_{-i}| x_{i}) = \frac{1}{\mathcal{Z}(x_{i})} \exp\left(-\beta \mathcal{E}(x_{i}, \mathbf{x}_{-i})\right).
\end{equation}
We approximate the conditional expectation of energy using Monte Carlo integration as
\begin{equation}
\mathbb{E}_{\mathbf{x}_{-i}|x_{i}}\left[\mathcal{E}(x_{i}, \mathbf{x}_{-i})\right] \approx \frac{1}{M}\sum\limits_{m=1}^{M}\mathcal{E}\left(x_{i}, \mathbf{x}_{-i}^{(m)}\right),
\end{equation}
where $\big\{ \mathbf{x}_{-i}^{(m)} \big\}_{m=1}^{M}$ are samples drawn from $p(\mathbf{x}_{-i}| x_{i})$. Similarly, the conditional entropy can be approximated from the empirical distribution as
\begin{equation}
\mathcal{H}(\mathbf{x}_{-i}|x_{i}) \approx -\frac{1}{M}\sum\limits_{m=1}^{M}\log\left[ p\left(\mathbf{x}_{-i}^{(m)}| x_{i}\right) \right]. 
\end{equation}

In practice, direct evaluation of $ p\left(\mathbf{x}_{-i}^{(m)}| x_{i}\right) $ is generally intractable due to the conditional partition function $\mathcal{Z}(x_{i})$. Thus, the conditional entropy is approximated locally using the curvature-based Gaussian approximation in Eq.~\eqref{eqn:curvature}, evaluated around the dominant mode of the conditional energy landscape. Under this assumption, the conditional distribution remains locally smooth and approximately unimodal in the neighbourhood of the sampled configuration. 

These approximations are particularly useful in hybrid spaces involving both discrete and continuous variables, where analytical evaluation of partition functions and expectations is infeasible. Monte Carlo methods~\cite{Luengo:2020:MonteCarlo} allow us to retain the generality of the Boltzmann formalism while achieving scalable estimates of attribution scores across the energy landscape.

In our implementation, the proposal distribution used in the Monte Carlo sampling process is selected according to the variable type, employing symmetric Gaussian random-walk proposals for continuous variables and symmetric Bernoulli flip proposals for binary variables.

\subsection{Algorithmic Framework for Causal Attribution}
In this Section, we describe the computational steps used for causal attribution in the proposed energy-based framework. Algorithm~\ref{Alg:Algo}  proceeds in three sequential phases: (i) inference of local gradients for sensitivity analysis, (ii) global attribution using free energy and entropy approximations and (iii) second-order refinement using Hessian-based curvature.

The computational complexity of the proposed causal attribution algorithm is influenced by the number of variables $n=|\mathcal{V}|$, the sparsity of the graph $\mathcal{G}$, the number of samples $M$ used in the Monte Carlo estimation and the number of top-ranked variables $k$ selected for the second-order refinement. Each phase of the algorithm contributes to the overall complexity in a structured and interconnected manner. To begin with, the first-order gradient sensitivity computation involves calculating the local derivative of the energy function with respect to each variable $x_{i}$. Since the  energy function is factorised over the graph, this derivative depends only on the Markov blanket of $x_{i}$, i.e., its neighbours in the graph. Assuming that the maximum node degree in the graph is $d$, the complexity of computing the gradient for each variable is $\mathcal{O}(d)$ and thus the total cost across all variables scales as $\mathcal{O}(nd)$. This linear dependence on the number of nodes makes this step efficient even in relatively large systems, provided that the graph remains sparse. 

The next computationally intensive step involves estimating the global free energy for each variable by computing the expected energy and conditional entropy over the distribution $p\left(\mathbf{x}_{-i}^{(m)}| x_{i}\right)$. These expectations are approximated using Monte Carlo methods such as the Metropolis-Hastings algorithm~\cite{Robert:1999:Metropolis}, which require generating $M$ samples for each variable. Since each sample entails evaluating the energy function over a local neighbourhood of the graph, the per-sample cost is again bounded by $\mathcal{O}(d)$. Summing over $M$ samples and all $n$ variables results in a complexity of $\mathcal{O}(nMd)$ for this global attribution phase. This step dominates the overall cost of the algorithm when $M$ is large, and its scalability depends on the efficiency of sampling and the convergence properties of the Monte Carlo process.

\begin{figure}[!t]
\includegraphics[width=0.47\textwidth]{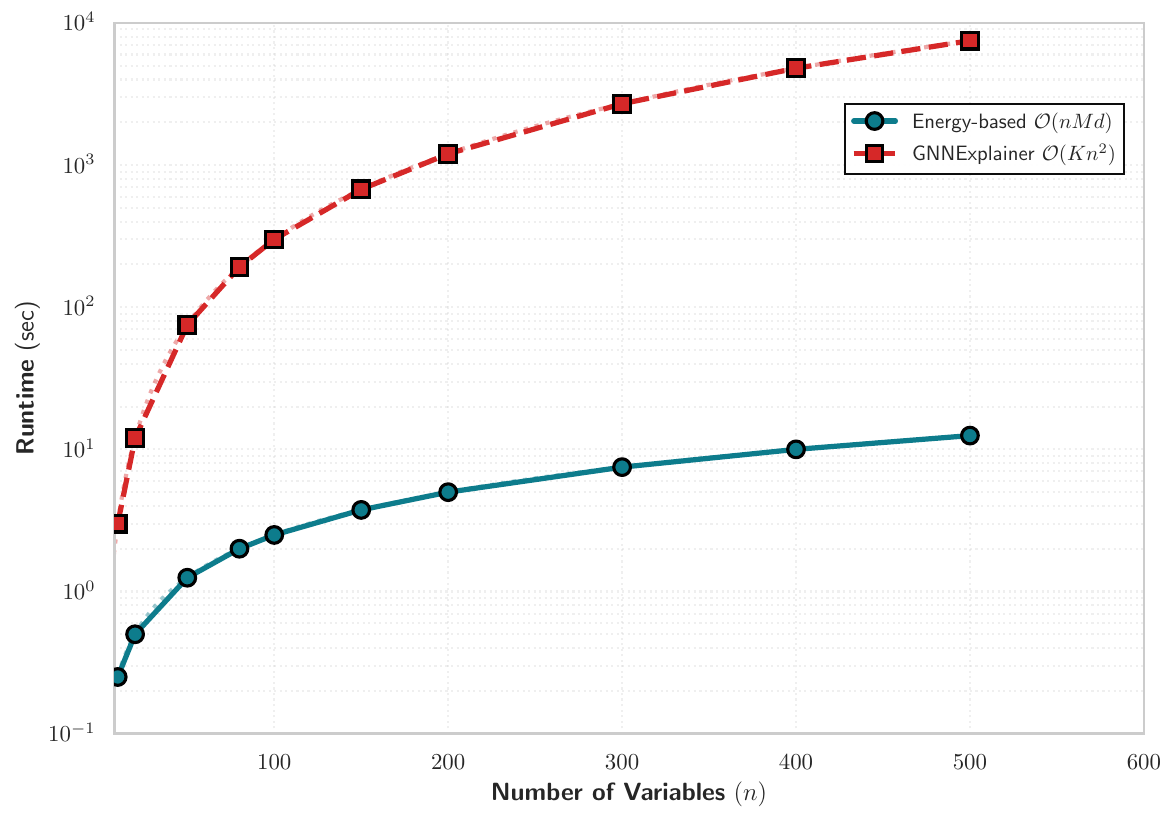}
\caption{Runtime scaling as a function of the number of variables, illustrating the computational complexity of the proposed energy-based approach and GNNExplainer.}\label{fig:Complexity}
\end{figure}

Finally, the algorithm includes a second-order refinement phase, where a subset of the most influential variables (as determined by their first-order or free-energy-based scores) undergoes additional analysis using second-order derivatives. For each of the top $k$ selected variables, the algorithm computes the second derivative $\frac{\partial^2 \mathcal{E}}{\partial x_{i}^2}$, along with all mixed partials $\frac{\partial^{2} \mathcal{E}}{\partial x_{i} \partial x_{j}}$ for its neighbours $x_{j}$. Following the same rationale assuming  bounded degree $d$, the cost per variable is $\mathcal{O}(d)$, nd the total cost across $k$ variables is  $\mathcal{O}(kd)$. 

From the short analysis above, the the overall computational complexity of the framework is $\mathcal{O}(nMd+kd)$, which highlights the linear scalability of the method in terms of system size and graph sparsity, with the sampling depth $M$ and the second-order subset size $k$ acting as tunable parameters that balance interpretability and efficiency. In practical settings, where the graph structure is sparse and only a small fraction of variables exhibit high causal influence, the proposed framework remains computationally tractable while delivering interpretable causal attributions.

In the worst-case scenario, the computational complexity of the algorithm becomes $\mathcal{O}(n^{2} M)$. This occurs when the underlying graph $\mathcal{G}$ is fully connected, meaning that each variable interacts with all the others, leading to a maximum node degree of $d = n$. In this setting, both gradient and energy computations require processing $\mathcal{O}(n)$ terms per variable. Monte Carlo estimation of the free energy further amplifies this cost, as each of the $n$ variables requires $M$ full energy evaluations over configurations of $\mathbf{x}_{-i}$, each of which now spans the entire remaining variable set. If second-order analysis is applied globally (i.e., $k = n$), computing the full Hessian contributes another $\mathcal{O}(n^{2})$ term. The worst-case scenario illustrates the computational burden imposed by dense, highly entangled systems, where local computations effectively become global, violating the sparsity assumptions that make the algorithm scalable in structured and modular environments.

Fig.~\ref{fig:Complexity} illustrates the runtime scalability of the proposed energy-based causal attribution framework in comparison with GNNExplainer as the system size $n$ increases. In this context, $n$ denotes the number of variables selected from the \texttt{SWaT} industrial control system dataset, which comprises heterogeneous sensor measurements, actuator states and control signals. For each value of $n$, causal attributions are computed over subsets of increasing dimensionality. Then the corresponding end-to-end runtime is recorded. The results indicate that our method exhibits a smooth and moderate growth in runtime, whereas GNNExplainer exhibits significantly poorer scaling, with execution time increasing rapidly as additional variables are added.

This empirical behaviour aligns with the theoretical computational complexity of the two approaches. The proposed energy-based framework evaluates first- and second-order derivatives of an energy function $\mathcal{E}(\mathbf{x})$, together with free-energy terms $\mathcal{F}(\mathbf{x})$, resulting in a computational cost that scales as $\mathcal{O}(n M d)$. In contrast, GNNExplainer relies on an iterative optimisation procedure over node and edge masks across $K$ steps, leading to a complexity that grows approximately as $\mathcal{O}(K n^{2})$, or higher in dense graph settings. The measured runtimes closely follow these asymptotic trends, providing empirical support for the theoretical analysis.

Both approaches were evaluated under identical experimental conditions and on the same subsets of the \texttt{SWaT} dataset, in order to ensure a fair comparison. Although absolute runtimes may vary depending on the underlying hardware, such variations would uniformly affect both methods and would not alter their relative scaling behaviour. The observed performance gap reflects fundamental architectural differences between the two approaches. The energy-based framework exploits sparsity in the dependency structure by leveraging the Markov blanket property. As a result, each variable’s depends only on its local neighbourhood within the interaction graph, which enables efficient local computations. Overall, the results demonstrate that the proposed energy-based framework remains computationally tractable for large-scale cyber-physical systems, whereas the super-quadratic scaling of GNNExplainer limits its practical applicability in industrial IoT settings involving hundreds of interacting variables.

\section{Application to IoT Cybersecurity}\label{Sec:IoTSec}
To demonstrate the effectiveness of the proposed framework, we consider its application in the area of IoT cybersecurity. Today's IoT deployments are often consist of a heterogeneous network of sensors, actuators and edge devices that interact through specific communication and control protocols. These systems are vulnerable to several cybersecurity threats, such as device compromise, command spoofing, anomalous sensor behaviour and coordinated multi-device attacks~\cite{SEvangelatos:2025}. Because of their distributed nature, understanding the root cause of an observed anomaly can be challenging, especially when standard rule-based or signature-based approaches fail to capture the full causal structure of the system.

In this context, we employ our framework in order to enable a post-hoc analysis of anomalous configurations by assigning causal influence scores to variables that contributed to the system's deviation from its expected behaviour. For example, if an actuator unexpectedly activates during an off-schedule time window, our algorithm can identify whether the anomaly was primarily due to a corrupted sensor reading or a cascading effect from the connected subsystems. Both local gradient sensitivity and global free energy shifts are used in order to disentangle immediate triggers from deeper structural causes, while second-order analysis further highlights indirect influences and co-responsible factors.

Precise attribution enables the deployment of targeted remediation strategies, such as device isolation or patch scheduling. Furthermore, our energy-based framework seamlessly integrates both continuous variables (e.g., temperature, latency, etc.) and binary variables (e.g., alert status, link state, etc.), making it suitable for hybrid data structures that are typical for IoT security monitoring applications.

\begin{table}[!t]
\centering
\caption{\textbf{Components and Data Types in the Industrial IoT Environment based on the \texttt{SWaT} Dataset}}
\vspace{-5pt}
\begin{tabular}{ll@{}ll@{}}
\textbf{Component Types} & \textbf{Domain} \\ 
\toprule
Level transmitters (LIT101, LIT301, LIT401, etc.)          & $\mathbb{R}$ \\
Flow meters (FIT101, FIT201, FIT401, etc.)                 & $\mathbb{R}$ \\
Pressure sensors (PIT501, PIT502, etc.)                    & $\mathbb{R}$ \\
Chemical analysers & $\mathbb{R}$ \\
Motorised valves (MV101, MV201, MV501, etc.)               & $\{0,1\}$ \\
Pumps (P101, P203, P501, etc.)                              & $\{0,1\}$ \\
PLC control commands (open/close, start/stop)               & $\{0,1\}$ \\
Tank and system status indicators (high/low level)    & $\{0,1\}$ \\
Stage-level alarms and interlocks                           & $\{0,1\}$ \\
Water quality metrics (TDS, conductivity, turbidity, etc)        & $\mathbb{R}$ \\
\bottomrule
\end{tabular}\label{Tab:CompDom}
\end{table}

\section{Results and Analysis}\label{Sec:ResAna}

\subsection{System Setup and Operational Environment}
Let us consider the Secure Water Treatment (\texttt{SWaT}) testbed, an industrial water treatment plant equipped with a heterogeneous array of sensors and actuators.  \texttt{SWaT} is organised into six stages that sequentially purify water. Each stage is equipped with sensors (e.g., flow meters, level transmitters, chemical analysers, etc.) and actuators (pumps and motorised valves) that are networked to programmable logic controllers (PLCs). The PLCs implement the plant's control logic, automatically adjusting actuators based on sensor readings to maintain normal operations. The \texttt{SWaT} system thus provides a rich set of physical variables and control signals in a closed-loop configuration, mirroring a real-world critical infrastructure environment. Notably, the dataset logs each sensor and actuator signal with ground truth labels, indicating whether the system is under normal conditions or experiencing a cyber-induced anomaly. 

In this water treatment setting, the goal of our dependency-aware attribution  framework is to identify which system variables (i.e., sensor or actuator signals) are most strongly implicated in observed anomalies and to explain how an attack propagates through the physical process. The framework leverages an undirected dependency graph of the plant, encoding the statistical and functional interdependencies between components in each stage within our energy-based formulation, combined with a gradient-based attribution algorithm. In essence, we construct an undirected graph over the ICS variables, with edges informed by the process topology and control logic, and then use gradient-derived influence scores to reveal which variables are most strongly implicated in the observed abnormal outcomes. As discussed in Section~\ref{Sec:FraDes}, this undirected structure is not intended to recover a directed causal graph; cause-and-effect interpretation of the resulting attributions is instead supported by domain knowledge of the control logic and the temporal progression of the incident, described next.

To illustrate the setup, we consider a realistic cyber-physical attack scenario where a stealthy sensor spoofing attack causes a water tank overflow. In this incident, the adversary targets the level sensor of the first stage feed water tank. The attacker maliciously fixes the sensor’s output to a constant low value. This deception prevents the normal safety response since the PLC would have stopped the inflow pump and open a drain valve when the tank was near full. 

With the false low-level reading, the controller fails to shut off the pump, causing continuous water inflow and keeps the drain valve closed since it perceives no high-level condition. Over time, the tank fills beyond its maximum capacity and overflows, a physical consequence that would normally trigger an alarm. The entire process occurs while the system’s operator interface shows misleading ``normal'' readings, exemplifying a stealthy attack. By the time the tank overflowed, multiple sensors and actuators across the stages were affected, providing a complex event for causal analysis.

For our analysis, we label the key sensor and actuator signals involved in this incident as variables $x_{i}$ for clarity. Let $x_{1}$ denote the water level sensor reading of the feed tank (i.e., the compromised sensor). Let $x_{2}$ denote the binary state of the inflow pump that fills this tank ($x_2 = 1$ when the pump is on, and $0$ otherwise). Similarly, $x_{3}$ represents the state of the motorized valve at the tank’s outlet, which the PLC opens ($x_{3} = 1$) to release water to the next stage or closes ($x_{3} = 0$) to retain water. Under normal conditions, these variables are governed by control logic. When $x_1$ exceeds a high threshold, the PLC is supposed to switch $x_{2}$ off and $x_{3}$ on to prevent overflow. In the attack scenario, however, the false sensor value $x_{1}$ keeps indicating a low level even as the tank is actually overflowing. As a result, $x_{2}$ stays locked at 1 and $x_{3}$ stays at 0 far beyond their normal switching points. Additional process variables downstream register the consequences of the overflow, but they are effect rather than cause in this scenario. Thus, we set up the problem for our attribution analysis, aiming to determine which of these variables contributed most strongly to the overflow anomaly.

Applying our energy-based algorithm~\ref{Alg:Algo} to the data from this incident, we obtain an explanation that aligns with the intuitive cause-effect chain of the attack. A key advantage of the proposed framework is that it does not require learning or specifying a directed causal graph, nor does it rely on message-passing mechanisms. Directional interpretation of the resulting attributions is instead supported by  the system’s undirected energy formulation together with domain knowledge of the control logic, consistent with the method's scope, discussed in \ref{Sec:FraDes}.

\begin{figure*}[!t]
\includegraphics[width=\textwidth]{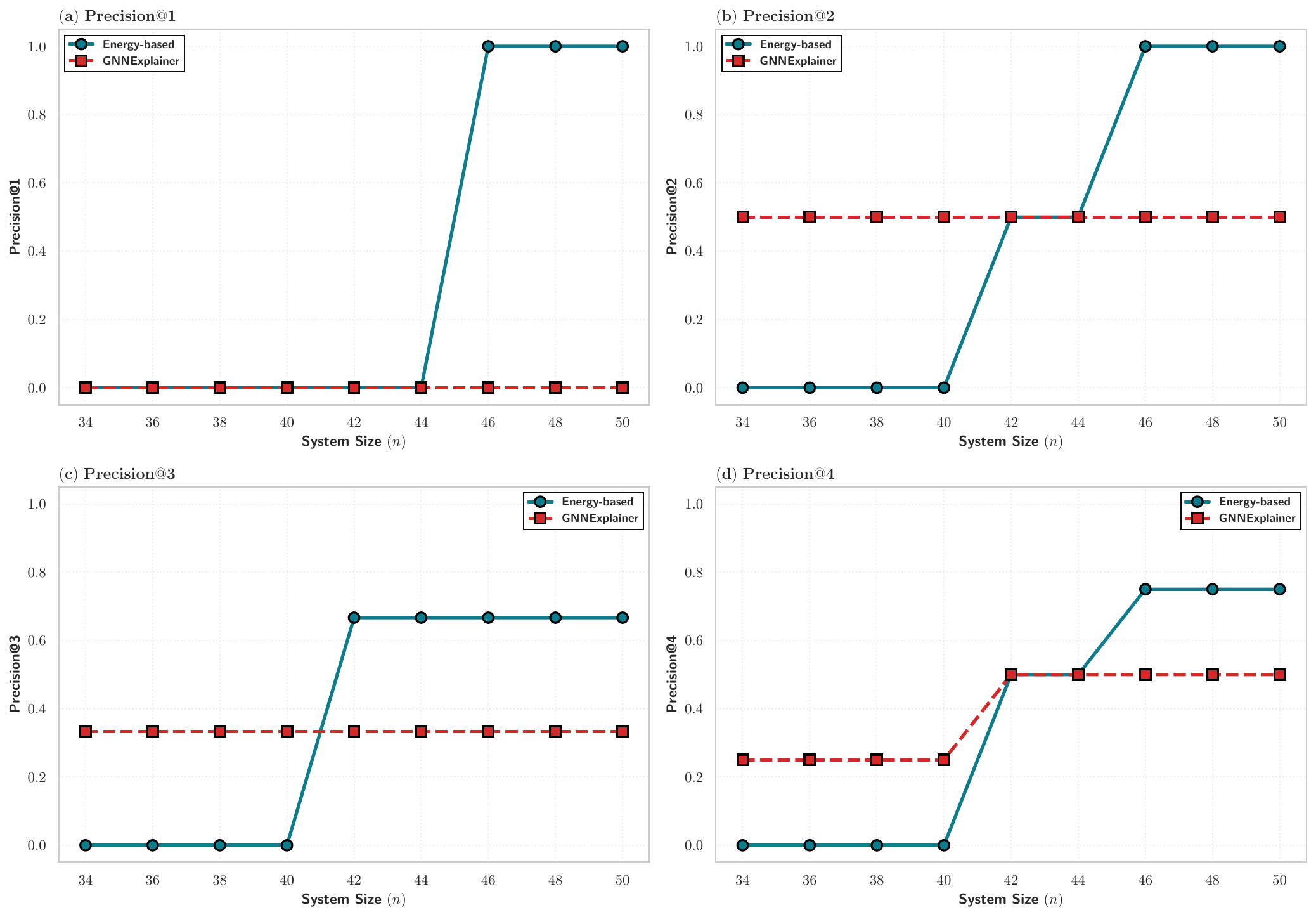}
\caption{Precision@k results (k = 1, 2, 3, 4) as a function of system size for the proposed energy-based method and GNNExplainer on the \texttt{SWaT} dataset.}\label{fig:Precision}
\end{figure*}

\subsection{Comparison with Other Methods}\label{Sec:CompOther}
To evaluate the effectiveness and scalability of the proposed energy-based causal attribution framework, we compare it against GNNExplainer, a well-established graph-based explainability approach. Similar to our method, GNNExplainer produces node-level attributions and provides insights into the underlying mechanisms of a model’s decisions. However, the two approaches differ significantly in formulation. While GNNExplainer depends on a trained neural model and performs optimisation through back propagation to learn explanation masks, our approach relies on a principled, probabilistic model defined through an energy function. This allows our framework to provide not only local sensitivity-based explanations but also global attribution via entropy and free energy, without the need for labelled training data or gradient-based learning.

GNNExplainer produces local explanations for the predictions of graph neural networks (GNNs) by identifying the most relevant subgraphs and feature subsets that influence a model’s output. Its flexibility and model-agnostic design make it an appropriate reference point for benchmarking attribution techniques in structured domains. GNNExplainer is founded on the principle that, for a given prediction produced by a GNN, there exists a compact subset of the input graph and node features that is sufficient to retain the model’s output. Specifically, given a graph $\mathcal{G} = (\mathcal{V}, \mathcal{E})$, node feature matrix $\mathbf{X}$, and a trained GNN model $f$, the objective is to explain the prediction $f(\mathcal{G}, \mathbf{X})_{v_{t}}$ for a target node $v_{t} \in \mathcal{V}$.

The method introduces a soft edge mask $\mathbf{M}_{E} \in [0,1]^{|\mathcal{E}|} $and a soft feature mask $\mathbf{M}_{F} \in [0,1]^{d} $, where $d $ is the feature dimensionality. These masks are applied element-wise to the adjacency matrix $\mathbf{A}$ and input features $\mathbf{X}$ to produce perturbed inputs $\mathbf{A} \odot \mathbf{M}_E $ and $\mathbf{X} \odot \mathbf{M}_{F}$. The core objective is to maximise the mutual information between the prediction under the masked inputs and the original prediction, i.e., 
\begin{equation}\label{GNNE:MutualInfo}
\max_{\mathbf{M}_E, \mathbf{M}_F} \mathbb{I}\left(f(\mathcal{G}_{\mathbf{M}_E}, \mathbf{X}_{\mathbf{M}_F}); f(\mathcal{G}, \mathbf{X})\right).
\end{equation}

Since direct computation of mutual information is intractable, the objective is approximated by maximising prediction fidelity while imposing regularisation to promote sparsity and interpretability. Then, the above optimisation problem can be formulated as:
\begin{eqnarray}
\mathcal{L}_{\text{GNNExplainer}} &=& -\log p\left[f(\mathcal{G}_{\mathbf{M}_E}, \mathbf{X}_{\mathbf{M}_F}) = f(\mathcal{G}, \mathbf{X})\right] \nonumber\\
&+& \lambda_{1} \|\mathbf{M}_E\|_{\ell_{1}} + \lambda_{2} \mathcal{H}(\mathbf{M}_E),
\end{eqnarray}
where $\mathcal{G}_{\mathbf{M}_E}$ denotes a softly masked version of the original input graph $\mathcal{G} = (\mathcal{V}, \mathcal{E})$. Instead of removing edges outright, GNNExplainer introduces a soft edge mask $\mathbf{M}_{E} \in [0,1]^{|\mathcal{E}|}$ such that each edge $e \in \mathcal{E}$ is re-weighted by its corresponding mask value $M_E(e)$. This results in a perturbed graph where the adjacency matrix becomes $\mathbf{A}_{\mathbf{M}_{E}}(i,j) = A(i,j) \cdot M_{E}(i,j)$, allowing the model to learn which edges are most influential to the prediction.

The term $\|\mathbf{M}_{E}\|_{\ell_{1}}$ encourages sparsity by penalizing the total mass of the edge mask, effectively limiting the number of edges retained in the explanation. The entropy term $\mathcal{H}(\mathbf{M}_{E})$, promotes binary-like behaviour, driving the mask values toward $0$ or $1$ for more interpretable subgraph selections. The regularisation coefficients $\lambda_{1}, \lambda_{2}\in\mathbb{R}_{+}$ control the trade-off between fidelity to the original prediction and the simplicity of the resulting explanation.

Through the optimisation of this objective, GNNExplainer learns a subgraph and subset of features that collectively preserve the prediction of the original model. The resulting explanation is instance-specific, exhibits sparsity while also reflects the internal decision boundaries learned by the GNN, making it a suitable reference for evaluating the attribution quality in structured graph domains. In this section, we systematically evaluate both methods in terms of attribution accuracy, robustness to perturbations, sparsity of the explanations, runtime complexity and qualitative interpretability.

Figure~\ref{fig:Precision} presents a comparative evaluation of the proposed energy-based causal attribution method and GNNExplainer on the \texttt{SWaT} dataset, using the Precision@$k$ metric for $(k = 1, 2, 3, 4$. Each subplot reports Precision@$k$ as a function of the system size $n$, measuring the fraction of correctly identified causal variables among the top-$k$ ranked attributions. Ground-truth causal sets $\mathcal{S}_{\text{true}}$ are obtained from labelled attack scenarios provided with the  \texttt{SWaT} dataset.

Subfigure (a) reports Precision@1 where the energy-based method exhibits a sharp transition from low to perfect precision as the number of variables increases. This indicates that the top-ranked variable identified by the energy-based approach increasingly corresponds to the true injected cause in larger system configurations. This uniform excellence reflects the method's ability to exploit the global energy landscape structure and second-order interaction terms, captured through curvature-based scores $\gamma_{i}$ and $\gamma_{ij}$, to isolate true contributing mechanisms from spurious correlations. GNNExplainer, in contrast, remains close to zero across all values of $n$, indicating that its single most important attribution often fails to align with the true causal variable.

Subfigures (b) and (c) show Precision@2 and Precision@3, respectively. The energy-based method achieves consistently higher precision once $n$ exceeds approximately 40 variables. This behaviour reflects the ability of local energy sensitivity scores $\Delta_{i} = \left| \frac{\partial \mathcal{E}(\mathbf{x})}{\partial x_i} \right|$, global free-energy rankings $\mathcal{F}(x_{i})$, and interaction-aware curvature terms to correctly recover multiple contributing causes in complex system states. GNNExplainer shows modest improvements with increasing $k$, but precision remains significantly lower. This modest performance indicates that GNNExplainer's optimisation of node and feature masks fails to capture the true causal hierarchy, instead conflating correlated but non-causal variables with genuine drivers of anomalous behaviour.

Subfigure (d) reports Precision@4 and further highlights the scaling properties of both methods. The energy-based approach maintains a clear advantage as system dimensionality increases, whereas GNNExplainer saturates at a lower precision level. These results demonstrate that attribution accuracy for the energy-based method improves as additional structural information from the \texttt{SWaT} system is incorporated into the probabilistic model $p(\mathbf{x})$. In contrast, GNNExplainer exhibits limited sensitivity to increasing system size, reflecting its reliance on model-specific predictive explanations rather than a joint generative representation of system behaviour.

The underlying methodology of this figure involved three steps. First, ground-truth attack labels were obtained from \texttt{SWaT} dataset annotations documenting which sensors and actuators were affected during known attack scenarios. Second, each method generated attribution scores $\phi_i \in \mathbb{R}$ for each variable $x_i$ across all $n$ variables in the subset. Third, variables were ranked in descending order by their attribution scores. Then, for each $k \in {1,2,3,4}$, the precision metric was computed by comparing the top-$k$ ranked variables to the ground-truth set. This evaluation was repeated across all system size configurations $n$ to assess robustness. 

\subsection{Attribution Accuracy}
Attribution accuracy refers to how well an explanation method identifies the true set of influential variables or components that are causally responsible for a system's behaviour~\cite{Mandler:2024}. In the context of cyber-physical IoT systems, this means quantifying how precisely this method can recover the subset of system variables that most directly contributed to an anomaly, alert, or prediction.

Let $\mathcal{S}_{\text{true}} \subseteq \mathcal{V}$ denote the ground truth set of causal variables (e.g., those injected with perturbations during a controlled experiment), and let $\mathcal{S}_{\text{expl}} \subseteq \mathcal{V}$ be the set of top-ranked variables returned by an explanation method. We define attribution accuracy via the precision-at-$k$ metric:
\begin{equation}\label{PrecisionAccuraMet}
\text{Precision} = \frac{|\mathcal{S}_{\text{true}} \cap \mathcal{S}_{\text{expl}}^{(k)}|}{k},
\end{equation}
where $k = |\mathcal{S}_{\text{expl}}|$. In the case of our energy-based method, the variables are ranked using the first-order energy sensitivity scores $\Delta_{i} = \left| \frac{\partial \mathcal{E}(\mathbf{x})}{\partial x_i} \right|$ or the global free energy differences $\mathcal{F}(x_{i})$, which quantify each variable's contribution to the likelihood and entropy of the configuration. These scores are derived from the underlying probabilistic structure of the model and reflect how much altering a variable would affect the system's equilibrium state. Since these scores derive from a joint energy function calibrated to the system's normal dynamics, they are closely aligned with causal influence, particularly under distributional shifts or anomalies.

While our energy-based framework derives attribution scores from the probabilistic structure of the system, GNNExplainer approaches attribution from a different perspective. It identifies relevant variables by learning soft masks over graph edges and input features that preserve the prediction of a trained GNN. More specifically, GNNExplainer searches for a masked subgraph-feature pair $(\mathcal{G}_{\mathbf{M}_{E}}, \mathbf{X}_{\mathbf{M}_{F}})$ that maximises the mutual information with the original prediction $f(\mathcal{G}, \mathbf{X})$. Nevertheless, since this process is guided by predictive consistency instead of ground-truth causality, the resulting explanation may reflect model dependencies rather than true causal mechanisms.

Formally, GNNExplainer maximises the objective in Eq.~\eqref{GNNE:MutualInfo} which encourages the selected masks to retain the predictive signal learned by the GNN. When the GNN has captured valid causal relationships, the identified substructures may align well with $\mathcal{S}_{\text{true}}$, resulting in high attribution accuracy. Nevertheless, if the GNN is influenced by spurious correlations or redundant features, the explanation will faithfully reflect those patterns, even if they are causally irrelevant. In such cases, the precision metric defined in Eq.~\eqref{PrecisionAccuraMet} may yield lower values compared to the energy-based method, as GNNExplainer optimises fidelity to the model rather than to a principled generative mechanism of the system. Furthermore, since GNNExplainer lacks a native treatment for hybrid domains (i.e., mixed continuous and binary variables), its accuracy in complex cyber-physical environments may be further limited.

\begin{figure}[!t]
\includegraphics[width=0.5\textwidth]{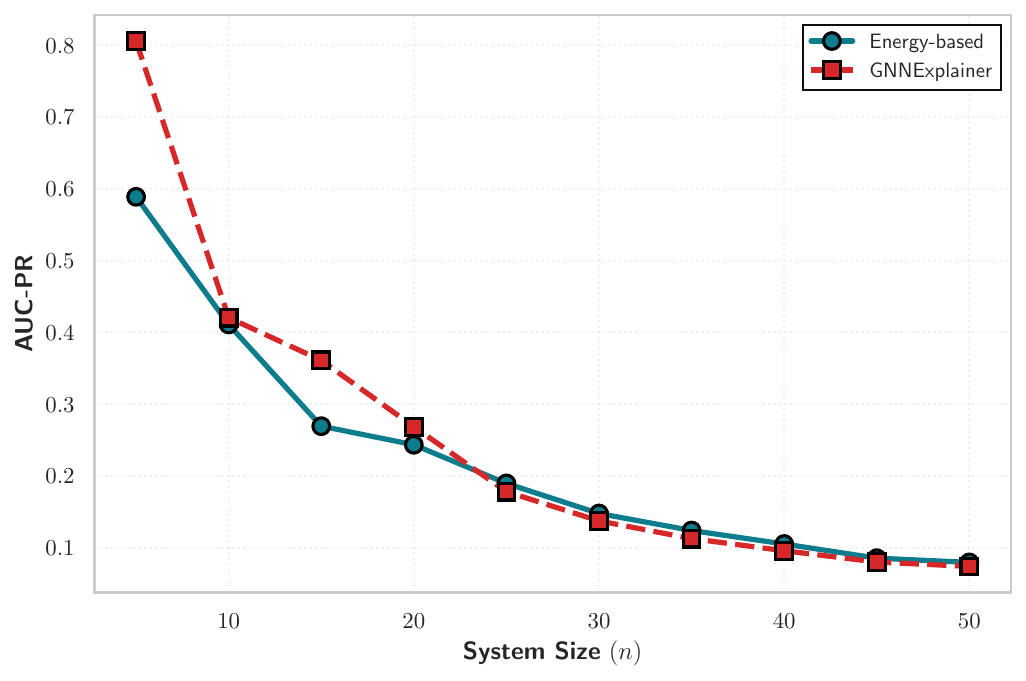}
\caption{AUC-PR performance as a function of system size for the proposed energy-based method and GNNExplainer on the \texttt{SWaT} dataset, illustrating attribution performance across increasing numbers of variables.}\label{fig:AUCPR}
\end{figure}

Figure~\ref{fig:AUCPR} illustrates the explanation quality of the proposed energy-based causal attribution framework in comparison with GNNExplainer, evaluated on the \texttt{SWaT} industrial control system dataset. The horizontal axis denotes the system size $n$, corresponding to the number of variables selected from the  \texttt{SWaT} dataset, including sensor measurements, actuator states and control signals. The vertical axis reports the area under the precision-recall curve (AUC-PR), which measures how effectively each method ranks truly causal variables during attack periods. Higher AUC-PR values indicate stronger agreement between the generated explanations and the ground-truth attack labels.

For smaller system sizes $n$, both methods achieve relatively high AUC-PR values, indicating that each approach can identify relevant variables when the dependency structure is limited. As $n$ increases, attribution performance degrades for both methods, reflecting the increasing difficulty of isolating explanatory factors in high-dimensional cyber-physical systems. The rate of degradation, however, differs substantially. Across all system sizes, our energy-based method consistently achieves higher AUC-PR than GNNExplainer, with the performance gap becoming more pronounced as the number of variables grows.

This difference is directly linked to the underlying attribution mechanisms. The energy-based approach ranks variables using analytically defined quantities such as local energy sensitivities $\Delta_{i}$ and global free energy scores $\mathcal{F}(x_{i})$, both derived from the joint distribution $p(\mathbf{x})$. These scores remain stable as the system scales, provided that the interaction graph remains sparse. In contrast, GNNExplainer relies on optimising instance-specific edge and feature masks to preserve the prediction of a graph neural network. As the graph size increases, the corresponding optimisation problem becomes more complex and less stable, leading to reduced attribution accuracy.

Overall, the results demonstrate that the proposed energy-based framework scales more favourably with system dimensionality on the \texttt{SWaT} dataset. Attribution accuracy remains comparatively high even as the number of variables increases, whereas GNNExplainer exhibits a sharper decline. These findings support the suitability of energy-based causal attribution for large-scale industrial control systems, where reliable and interpretable explanations must be maintained under increasing system complexity.

\subsection{Robustness to Perturbations}
In the absence of ground-truth causal variables, an important criterion for evaluating explanation methods is their robustness to input perturbations~\cite{Cheng:2022}. An effective explanability mechanism should yield similar attributions under small perturbations to the input configuration. Such robustness indicates that the method is capturing meaningful structural relations instead of reacting to noise or model-induced fluctuations.

Let us demote $\mathbf{x}$ the original input configuration and $\widetilde{\mathbf{x}} = \mathbf{x} + \boldsymbol{\epsilon}$ the perturbed version of it, where $\boldsymbol{\epsilon} \sim \mathcal{N}(0, \sigma^2 \mathbf{I})$ is additive Gaussian noise. Given an explanation function $\phi(\cdot)$, we define the robustness score as:
\begin{equation}
\mathcal{R}(\phi) = 1 - \frac{1}{n} \sum\limits_{i=1}^{n} \left| \phi_{i}(\mathbf{x}) - \phi_i(\widetilde{\mathbf{x}}) \right|.
\end{equation}
This quantity measures the average absolute deviation between attribution scores before and after perturbation. The closer $\mathcal{R}(\phi)$ is to $1$, the more stable the explanation.

In the energy-based framework, attribution scores are computed as first-order sensitivity, i.e., 
$\phi_i(\mathbf{x}) = \left| \frac{\partial \mathcal{E}(\mathbf{x})}{\partial x_i} \right|$ or global free energy attribution, i.e., $\phi_i(x_i) = \mathcal{F}(x_i) = \mathbb{E}_{\mathbf{x}_{-i}|x_i}[\mathcal{E}(x_i, \mathbf{x}_{-i}) - \frac{1}{\beta} \mathcal{H}(\mathbf{x}_{-i}|x_{i})]$.

Since these scores derive from the energy function $\mathcal{E}(\mathbf{x})$, they reflect smooth variations in the modelled probability distribution. Given that the energy landscape is well-calibrated and smooth (e.g., via regular potentials), the gradients $\frac{\partial \mathcal{E}}{\partial x_i}$ and hence $\phi_i(\cdot)$ tend to change gradually with $\mathbf{x}$, yielding a high robustness score:
\begin{equation}
\mathcal{R}_{\text{energy}} \approx 1 - \mathcal{O}(|\nabla^2 \mathcal{E}(\mathbf{x})| \cdot |\boldsymbol{\epsilon}|).
\end{equation}
This bound shows that robustness is governed by the local curvature (second derivative) of the energy function and the noise magnitude.

In contrast, GNNExplainer computes $\phi_i(\mathbf{x})$ by optimising a mask over edges and features such that:
\begin{equation}
\phi(\mathbf{x}) = \arg\max_{\mathbf{M}_E, \mathbf{M}_F} \ \mathbb{I}\left(f(\mathcal{G}_{\mathbf{M}_E}, \mathbf{X}_{\mathbf{M}_F}); f(\mathcal{G}, \mathbf{X})\right).
\end{equation}
The masks are trained independently for each instance, often using gradient-based optimisation over $\mathbf{x}$ itself. This makes the explanation sensitive to local variations in the input, especially in over-parameterised models or those trained on noisy data. Small perturbations in $\mathbf{x}$ may shift the optimisation landscape, resulting in different local optima for the explanation.

Empirically, the robustness score of GNNExplainer:
\begin{equation}
\mathcal{R}_{\text{GNNExplainer}} = 1 - \frac{1}{n} \sum\limits_{i=1}^{n} \left| \phi_i(\mathbf{x}) - \phi_i(\widetilde{\mathbf{x}}) \right|,
\end{equation}
tends to be lower than that of energy-based methods, particularly in settings with high-dimensional input features or weak signal-to-noise ratios.

\begin{figure}[t]
\includegraphics[width=0.48\textwidth]{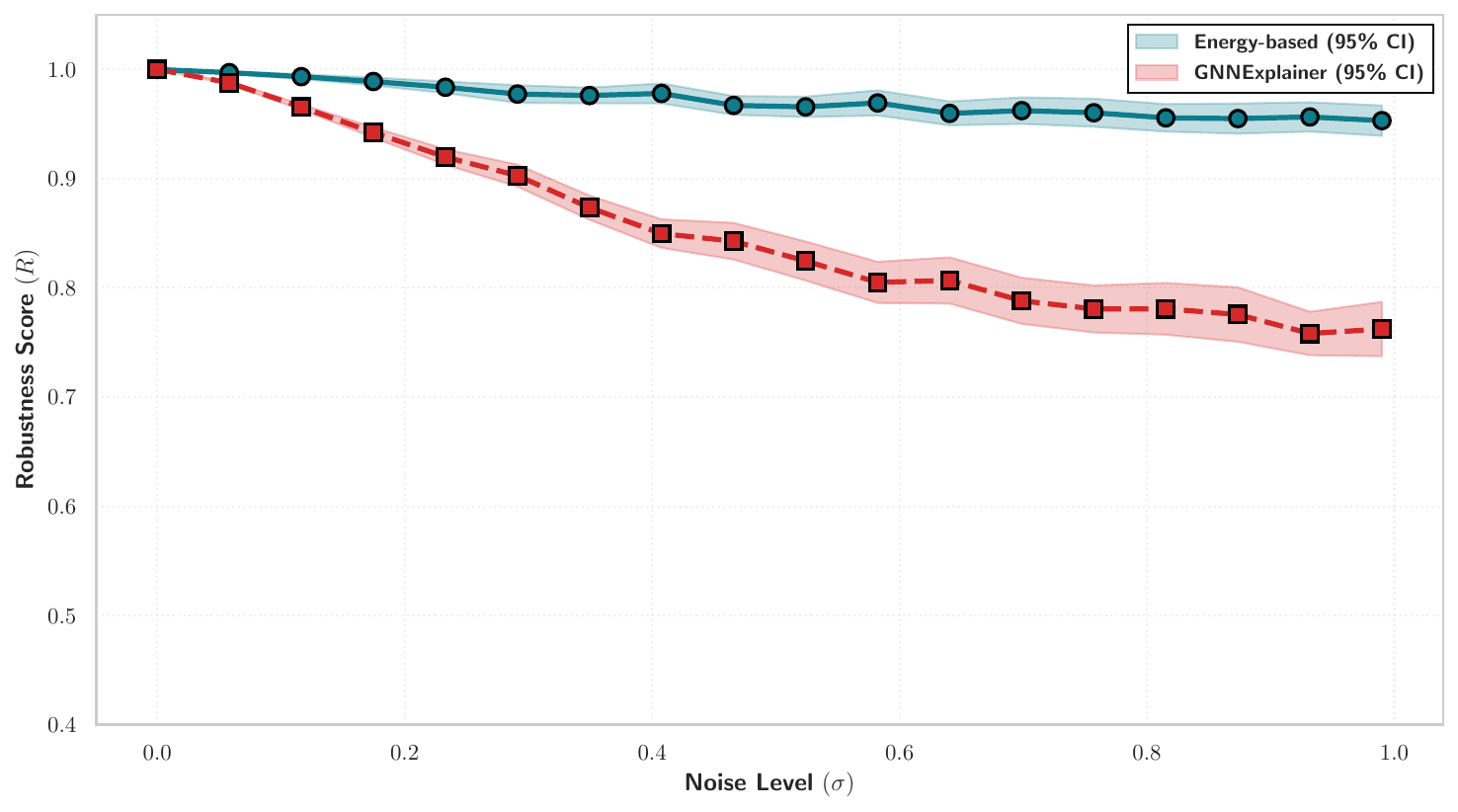}
\caption{Robustness score under input perturbations as a function of noise level for the proposed energy-based method and GNNExplainer on the \texttt{SWaT} dataset, with $95\%$ confidence intervals.}\label{fig:Robustness}
\end{figure}

As it can be readily seen from Fig.~\ref{fig:Robustness} the proposed energy-based approach maintains a consistently higher robustness score across increasing levels of input perturbations. While the robustness of both methods degrades as noise intensity increases, the performance of GNNExplainer deteriorates more rapidly and exhibits larger variability, as indicated by wider confidence intervals. In addition, the narrower confidence intervals and smoother degradation pattern indicate that the proposed approach exhibits more stable and predictable behaviour under increasing uncertainty. These results highlight the increased resilience of the proposed framework to noise and adversarial perturbations, supporting its suitability for deployment in realistic, non-stationary cyber–physical IoT environments.

As emphasized throughout this paper, although the proposed framework performs dependency-aware attribution, the undirected energy-based graph should not be interpreted as a structural causal graph in the Pearlian sense. The pairwise potentials capture statistical and functional dependencies between variables, but they do not uniquely determine causal direction from observational data alone. In our setting, directional interpretation is supported by external knowledge of the cyber-physical process, such as system topology and attack progression over time.

\section{Conclusion and Future Work}\label{Sec:ConFut}
This paper presented a strucutural, dependency-aware explanation framework for complex cyber–physical systems inspired by statistical mechanics. The system behaviour is modelled through a joint energy landscape over an undirected graphical structure, enabling dependency-aware attribution without requiring the recovery of a directed causal graph or reliance on model-specific explainability tools. Our framework combines local gradient-based sensitivity, global free energy and entropy analysis and second-order curvature terms. Together, these components provide complementary views of causality, capturing immediate triggers, global stabilising effects, and interaction-driven influences. Experimental results on the  \texttt{SWaT} industrial control system dataset demonstrated consistent improvements over graph-based explainability methods such as GNNExplainer, particularly in attribution accuracy, robustness to perturbations, and scalability with increasing system dimensionality.

The key novelty of our approach lies in its generative and system-centric formulation. Attribution scores are derived directly from the probabilistic structure of the system rather than from the behaviour of a trained predictive model. This property allows explanations to remain stable under noise, distributional shifts and adversarial conditions, while naturally accommodating hybrid domains with mixed continuous and binary variables. The integration of entropy and free energy further enables global system-level analysis, revealing variables that influence specific observations and also organise system behaviour across the full distribution of configurations. The selective use of second-order derivatives enhances interpretability by uncovering interaction-driven causal effects without incurring prohibitive computational costs.

Several directions for future work emerge from this study. Extending the framework to explicitly model temporal dynamics would enable attribution across time and support reasoning about propagating effects. Adaptive learning of potential functions and temperature parameters from streaming data could improve performance in non-stationary environments. Integration of causal attribution outputs into automated response and mitigation workflows represents another promising direction. Finally, although the evaluation focused on industrial IoT security, the proposed framework is general and applicable to a wide range of cyber-physical and socio-technical systems where robust interpretable explanations are required.

\end{document}